\documentclass[letterpaper, 10 pt, conference]{ieeeconf}  

\IEEEoverridecommandlockouts                              

\usepackage{amsmath,amssymb} 
\usepackage{color}
\usepackage{graphicx}

\definecolor{multi_trans}{RGB}{192,31,219}
\definecolor{pedformer}{RGB}{243,144,11}
\definecolor{sep_dec}{RGB}{19,76,227}
\definecolor{no_int}{RGB}{22,222,232}
\definecolor{gt}{RGB}{16,244,15}

\usepackage{pifont}
\newcommand{\cmark}{\ding{51}}%
\newcommand{\xmark}{\ding{55}}%

\usepackage{hyperref}       
\usepackage{multirow}
\graphicspath{{images/}}
\usepackage{subfig}
\usepackage{array}
\newcolumntype{?}{!{\vrule width 1pt}}

\title{\LARGE \bf
PedFormer: Pedestrian Behavior Prediction via Cross-Modal Attention Modulation and Gated Multitask Learning
}

\author{Amir Rasouli*$^{1}$ and Iuliia Kotseruba$^{12}$
\thanks{*Corresponding author {\tt\small amir.rasouli@huawei.com}}
\thanks{$^{1}$Noah's Ark Laboratory,  Huawei, Canada.
        }%
\thanks{$^{2}$Department of Electrical Engineering and Computer Science, York University,
        Toronto, Canada
        }
}
\begin{document}

\maketitle
\thispagestyle{empty}
\pagestyle{empty}

\begin{abstract}
Predicting pedestrian behavior is a crucial task for intelligent driving systems. Accurate predictions require a deep understanding of various contextual elements that potentially impact the way pedestrians behave. To address this challenge, we propose a novel framework that relies on different data modalities to predict future trajectories and crossing actions of pedestrians from an ego-centric perspective. Specifically, our model utilizes a cross-modal Transformer architecture to capture dependencies between different data types. The output of the Transformer is augmented with representations of interactions between pedestrians and other traffic agents conditioned on the pedestrian and ego-vehicle dynamics that are generated via a semantic attentive interaction module. Lastly, the context encodings are fed into a multi-stream decoder framework using a gated-shared network. We evaluate our algorithm on public pedestrian behavior benchmarks, PIE and JAAD, and show that our model improves state-of-the-art in trajectory and action prediction by up to $22\%$ and $13\%$ respectively on various metrics. The advantages brought by components of our model are investigated via extensive ablation studies.
\end{abstract}

\section{Introduction}
Pedestrian behavior prediction is a fundamental task for intelligent driving in urban environments. What makes it challenging is that pedestrians' actions are influenced by the interplay of various environmental factors, such as social interactions and traffic dynamics \cite{Rasouli_2019_ITS, Rasouli_2017_IV}. Past approaches investigated two different ways of predicting pedestrian behavior: future trajectories \cite{Shi_2021_CVPR,Rasouli_2019_ICCV} and forthcoming actions, e.g. crossing the road \cite{Liu_2020_RAL,Rasouli_2019_BMVC}. However, recent studies \cite{Rasouli_2021_ICCV,Liang_2019_CVPR} showed that these modes of prediction are complementary. For example, a pedestrian expecting to move from one side of the road to another implies a future crossing action. 

In this paper, we exploit the complementarity of trajectory and action predictions in a multitask learning framework. Our approach relies on a cross-modal attention modulation technique to encode pedestrians' and scene dynamics using multimodal data. This encoding is augmented with representations of interactions between pedestrians and their surroundings captured by an attentive interaction module that uses semantic scene information as well as pedestrians' and the ego-vehicle dynamics. The combination of interaction representation and modality encodings is fed into a hybrid-gated multi-stream decoder to predict future trajectories and crossing actions. Evaluations on pedestrian behavior benchmark datasets, PIE \cite{Rasouli_2019_ICCV} and JAAD \cite{Rasouli_2017_ICCVW}, demonstrate that the proposed method significantly improves state-of-the-art performance on both tasks.

\section{Related Works}
\subsection{Pedestrian Behavior Prediction}
Human behavior prediction has many applications in different domains \cite{Liu_2020_ECCV,Piergiovanni_2020_ECCV,Joo_2019_CVPR,Yao_2018_CVPR,Epstein_2020_CVPR,Qi_2020_CVPR,Felsen_2017_ICCV}. In traffic scene understanding, behavior prediction either takes the form of trajectory or crossing action prediction.

\subsubsection{Trajectory Prediction} 
\sloppy Pedestrian trajectory prediction algorithms take two forms. The dominant approach is making predictions from bird's eye view surveillance-like videos recorded with a fixed camera \cite{Dendorfer_2021_ICCV,Shafiee_2021_CVPR,Hu_2020_CVPR,Mohamed_2020_CVPR,Sun_2020_CVPR,Sun_2020_CVPR_2,Mangalam_2020_ECCV,Choi_2019_ICCV,Zhang_2019_CVPR,Sadeghian_2019_CVPR,Gupta_2018_CVPR}. 

Another group of algorithms makes predictions from images captured with a moving monocular camera \cite{Neumann_2021_CVPR,Makansi_2020_CVPR,Malla_2020_CVPR,Rasouli_2019_ICCV,Yagi_2018_CVPR,Yao_2019_ICRA,Yao_2019_IROS,Bhattacharyya_2018_CVPR,Chandra_2019_CVPR}. However, lack of depth and global positions of agents and camera ego-motion make predictions quite challenging in this context. To overcome these difficulties, ego-centric models rely on multimodal sources of data to reason about the future trajectories of pedestrians. For instance, the authors of \cite{Bhattacharyya_2018_CVPR} employ a two-stream encoder-decoder architecture. One stream predicts ego-vehicle motion by relying on observed ego-motion and visual context, and the second stream uses encoded observed pedestrian trajectories and predicted ego-motion to forecast future trajectories. Other multi-stream models use pedestrian action, e.g. waiting to cross \cite{Malla_2020_CVPR},  or intention of performing an action, e.g. crossing the road \cite{Rasouli_2019_ICCV}.  In \cite{Yagi_2018_CVPR}, the authors use pedestrian poses in addition to trajectories and ego-motion. Their model has three streams of 1D convolutional layers that encode different data modalities. The outputs of the encoders are combined and fed into a convolutional decoder to generate final predictions. 
 
\subsubsection{Action Prediction}
In the driving context, predicting crossing actions of pedestrians is helpful for assessing risk and anticipating pedestrian trajectories for safe motion planning  \cite{Kotseruba_2021_WACV,Chaabane_2020_WACV,Saleh_2019_ICRA,Gujjar_2019_ICRA,Rasouli_2017_ICCVW,Liu_2020_RAL,Rasouli_2019_BMVC,Aliakbarian_2018_ACCV}. To achieve this goal, it is common to use streams of images. In some approaches \cite{Gujjar_2019_ICRA,Chaabane_2020_WACV}, the images are used in a generative architecture to produce future scenes, which in turn are classified as containing crossing and non-crossing events. Alternatively, in \cite{Saleh_2019_ICRA} the authors first localize pedestrians and then use a 3D convolutional architecture to classify their actions in the future frames. 

Similar to ego-centric trajectory prediction algorithms, many crossing action prediction models use multimodal data \cite{Liu_2020_RAL,Rasouli_2019_BMVC,Aliakbarian_2018_ACCV,Rasouli_2017_ICCVW}. For example, the authors of \cite{Rasouli_2019_BMVC} use visual appearance of pedestrians and their surroundings, their locations, poses, and ego-vehicle speed in a hierarchical architecture. The model in \cite{Aliakbarian_2018_ACCV} relies on visual features of the scenes, optical flow, and ego-vehicle dynamics each of which are processed with individual recurrent networks, combined and fed to another recurrent network for final prediction. The approach of \cite{Kotseruba_2021_WACV} uses a hybrid architecture in which visual features are encoded using 3D convolutional layers in addition to other modalities, such as poses, trajectories, and ego-speed, that are encoded using recurrent networks. The encodings are combined using self-attention modules before the final prediction.

\subsubsection{Multitask Prediction}
\sloppy Multitask learning has a wide range of applications in various computer vision domains from behavior understanding \cite{Luvizon_2018_CVPR,Guo_2018_ECCV,Hu_2018_ECCV,Du_2019_CVPR} and object recognition \cite{Mallya_2018_ECCV,Zeng_2019_ICCV,Tang_2019_ICCV,Hassani_2019_ICCV} to intelligent driving \cite{Rasouli_2021_ICCV,Casas_2018_CORL,Kendall_2018_CVPR,Liang_2019_CVPR_2,Liu_2019_CVPR,Wu_2020_CVPR_2}. These works emphasize the effectiveness of multitask learning for improving performance. 

Multitask learning is also widely used in the context of pedestrian behavior prediction \cite{Hasan_2018_CVPR,Fernando_2018_ACCV,Liang_2019_CVPR,Zhang_2020_CVPR}. The model in \cite{Hasan_2018_CVPR} predicts trajectories and head poses of pedestrians simultaneously and shows that correlation between the tasks can result in improved trajectory forecasting.  The method in \cite{Liang_2019_CVPR} predicts future activities of pedestrians, e.g. interacting with a car, as an auxiliary task to improve trajectory prediction using a recurrent framework. 

In an ego-centric setting, the method in \cite{Rasouli_2021_ICCV} simultaneously predicts pedestrian crossing actions and trajectories and shows significant improvements on both tasks due to their complementary roles. The proposed bifold approach decodes contextual information using individual and shared decoders whose outputs are averaged to make final prediction. This approach, however, can be computationally inefficient as two sets of predictions take place. In addition, using separate decoders can negatively impact the model's ability to capture effects that prediction tasks have on each other.   

The method we propose is a multitask learning framework similar to \cite{Rasouli_2021_ICCV} for predicting future pedestrian trajectories and crossing actions from an ego-centric perspective. Differently from the past work, we propose a unified decoding mechanism that uses a gated flow mechanism between shared and individual task decoders. The predictions are made once using the outputs of the task decoders. 

\subsection{Multimodal Data}
To compensate for the lack of global position information, many ego-centric behavior prediction methods rely on multimodal data. The common approach is to process different modalities individually and then collapse them into a single representation by addition or concatenation \cite{Rasouli_2019_ICCV,Liang_2019_CVPR} or using attention to reweigh them prior to merging \cite{Kotseruba_2021_WACV}. Alternatively, in \cite{Rasouli_2019_BMVC}, representation are organized hierarchically according to their complexity level (visual at the bottom and ego-speed at the top) and merged gradually, whereas in \cite{Choi_2021_CVPR}, modality representations are projected into a shared latent space prior to prediction.

Independently encoding different modality representations makes them more susceptible to noise and makes capturing step-wise correlation between them difficult. The model in \cite{Rasouli_2021_ICCV} addresses these issues with a bifold mechanism which consists of individual encoders and a shared encoder that receives an embedded combination of all input data. The outputs of all encoders are then combined for the final prediction. Although effective, this approach cannot capture the mutual influences between data modalities and does not have any mechanisms to control the weights of different data types in shared representations. To this end, we propose an attention mechanism in which different data modalities are combined pair-wise via multi-head attention units the outputs of which are combined and fed into a Transformer architecture to produce the final input encoding.

\begin{figure*}[!ht]
\vspace{0.2cm}
\centering
\includegraphics[width=0.8\textwidth]{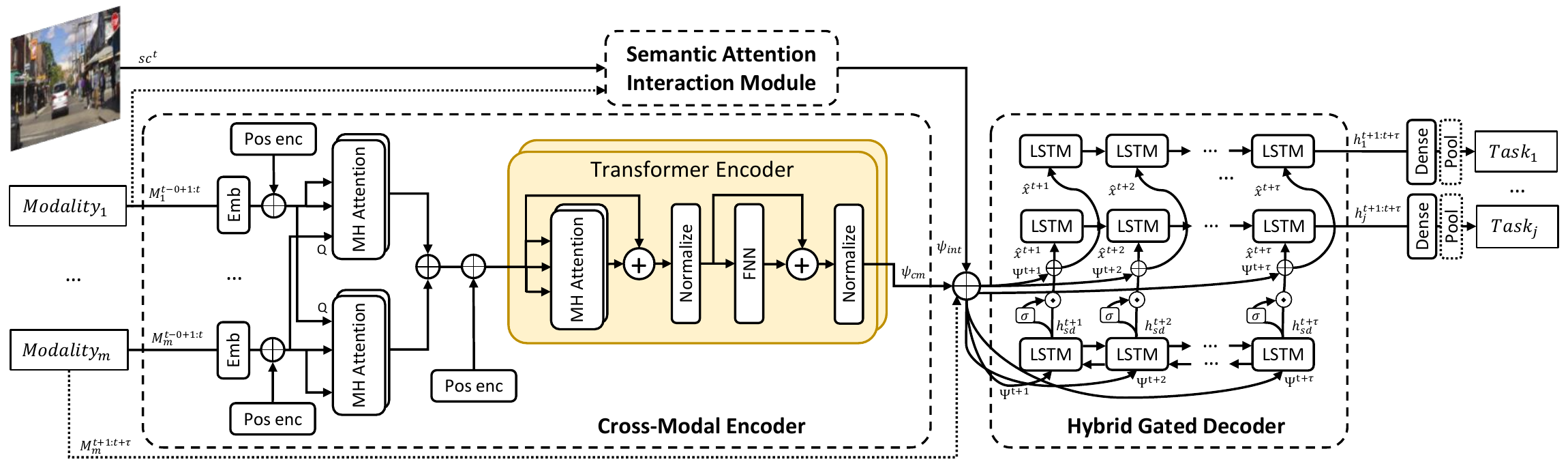}
\caption{Proposed architecture comprised of three modules: 1) A \textit{cross-modal encoder} which uses multi-head attention to capture pair-wise dependencies between data modalities, 2) \textit{semantic attention interaction module} (SAIM) that encodes interactions between ego-vehicle, environment, and other agents, and 3) a \textit{hybrid gated decoder} which produces predictions from a combination of modality encodings and interaction representation.  $\oplus$ denotes concatenation operation.}
\vspace{-2em}
\label{main_diagram}
\end{figure*}
\vspace{-0.2cm}
\subsection{Pedestrian Interaction with Environment}

A key challenge in multi-agent behavior prediction task is modeling interactions between agents and their surroundings. A common approach for pedestrian behavior prediction is to use pedestrians' positions and determine their effect on each other in the future. This can be done either implicitly by co-processing trajectories within a neighboring region using methods, such as social pooling \cite{Gupta_2018_CVPR,Sun_2020_CVPR_2,Mangalam_2020_ECCV} or more explicitly by determining the importance of interacting trajectories using attention-based \cite{Sadeghian_2019_CVPR,Zhang_2019_CVPR,Park_2020_ECCV} or graph architectures \cite{Yau_2021_ICRA,Li_2020_NeurIPS,Mohamed_2020_CVPR,Sun_2020_CVPR,Kosaraju_2019_NeurIPS,Kipf_2018_ICML}.

Although position-based techniques have been used in ego-centric context \cite{Malla_2020_CVPR,Chandra_2019_CVPR}, their utility is limited if global positions of agents are not available. In the ego-centric view, image coordinates of agents can be easily confused. For example, two pedestrians with different heights that are moving next to one another and two pedestrians of the same height but moving at different distances from the ego-vehicle can yield similar trajectories. To this end, the authors of \cite{Liu_2020_RAL} use visual encodings of the agents in addition to their pair-wise spatial locations in the form of union of the bounding boxes surrounding them. The method in \cite{Rasouli_2021_ICCV} removes the dependency on spatial information by relying only on semantic parsing of the scenes. Here, the semantic map of the scene is grouped by object categories, which are processed individually and combined via an attention module. Both approaches have several issues in common. First is the scalability, as the former method requires pair-wise reasoning for all agents and the latter involves visual processing for each pedestrian and surrounding people. Second, neither approach accounts for ego-vehicle motion when reasoning about the interactions. Finally, use of convolutional networks for processing scene information can introduce inductive biases such as location invariance which is not desirable when reasoning about agents' positions. We address these issues by proposing a novel method that is based on visual attention networks in which semantic information about the scene is divided into channels containing objects with shared characteristics. The semantic map is then divided into local patches, which after a series of embedding operations are fed into a multi-head attention network followed by a global attention module that produces a summary given pedestrian and ego-vehicle dynamics.

\noindent\textbf{Contributions:} 1) We propose a novel pedestrian behavior prediction model that simultaneously predicts pedestrian trajectory and crossing action in ego-centric video sequences. Our model employs a cross-modal attention-based encoding architecture for processing multimodal data, an interaction modeling module that uses visual attention architecture to encode interactions between pedestrians and their surroundings, and a hybrid gated decoder mechanism for processing context representation. 2) We evaluate the model on two publicly available benchmarks datasets and show that it significantly improves state-of-the-art performance on both tasks. 3) We conduct ablative studies to highlight the contributions of different modules to the overall performance.

\vspace{-0.1cm}
\section{Method}
\vspace{-0.1cm}
\subsection{Problem Formulation}
Behavior prediction is formulated as a multi-objective optimization problem. The goal is to learn the following distribution $p(Tasks| Modalities)$ where $Tasks = \{l^{t+1:t+\tau}_i, a_i, dl^{t+\tau}_i\}$ correspond to trajectory, future crossing action and discrete location respectively for some pedestrian $1 < i < n$. $Modalities = \{l^{t-o+1:t}_i, dl^{t-o+1:t}_i, sc^{t}, em^{t-o+1:t+\tau}\}$ denote $i$th pedestrian's observed coordinates and discrete location, and scene and ego-vehicle motion respectively. We denote the length of observations as $o$ and prediction as $\tau$.

\subsection{Architecture}
The proposed architecture shown in Figure \ref{main_diagram} has three modules: \textit{cross-modal encoder} that captures pair-wise dependencies between data modalities via multi-head attention, \textit{semantic attention interaction module (SAIM)} that leverages semantic information to model interactions between pedestrians and their surroundings, and \textit{hybrid gated decoder} that controls flow of information between shared and independent decoders for final prediction. Implementation details for each module are given in the following subsections.
\subsection{Cross-modal Encoder}
\noindent
\textbf{Input data modalities:}
\textit{Pedestrian location and velocity}. \sloppy We use normalized 2D coordinates of pedestrians in the image plane $[(x_1,y_1), (x_2,y_2)]$, corresponding to top-left and bottom-right corners of the pedestrian bounding boxes. This allows the model to localize the pedestrian with respect to the ego-vehicle and other scene elements. Pedestrian velocity at time $t$ is computed as difference between their 2D positions at time $t$ and $t-1$.
 
\textit{Discrete locations.} Following \cite{Rasouli_2021_ICCV}, we transform pedestrian coordinates by dividing the image plane into a grid with $N \times M$ cells with unique class labels. At each time step, the cell whose center is closest to the pedestrian is selected.

\textit{Ego-vehicle motion} captures both the changes in the scale of pedestrians in ego-centric view and impact of the ego-vehicle motion on pedestrian behavior. We use three sources of information to characterize ego-motion: vehicle speed $s$ and vehicle velocity along $x$ and $z$ axes denoted as $v_x$ and $v_z$, respectively. 

\textbf{Cross-modal attention modulation.} It is important to capture correlations between different data modalities to generate representations robust to noise or missing data \cite{Ruder_2017_arxiv}. Bifold method proposed in \cite{Rasouli_2021_ICCV} processes multimodal data using hard parameter sharing and individually, and combines the results for final representation. However, this method is problematic when pair-wise correlations across different modalities are not similar.

We propose to instead use multi-head attention units, $\operatorname{Attn_{MH}}$. For each of $m$ data modalities we define $m-1$ attention units as follows,
\begin{equation}
    \begin{gathered}
\operatorname{Attn_{MH}}(Q_j,K_m,V_m)= (head_1 \oplus head_2 \dots head_{k})W^O\\
head_k = \operatorname{S-Attn}(Q_jW_k^Q,K_mW_k^K,V_mW_k^V)\\
\operatorname{S-Attn}(Q,K,V) = Softmax(QK^{\top})V,
\end{gathered}
\end{equation}

\noindent where $K_m$ and $V_m$ are key and value for $m$th data modality and $Q_j$ is the query based on  $j$th data modality, where $j\neq m$. $\operatorname{S-Attn}$ denotes self-attention network with $k$ heads and $\oplus$ denotes concatenation operation. Prior to entering $\operatorname{Attn_{MH}}$ units, data from each modality is fed into an embedding layer followed by concatenation with a positional encoding.

\textbf{Transformer encoder.} Final encoding is generated using a Transformer architecture \cite{Vaswani_2017_NIPS} which receives the concatenated output of multi-head attention units and a positional encoding as input. The encoder consists of $l$ identical layers connected sequentially, i.e. the output of one layer is the input of the next one. The first module in each layer is self-attention followed by a feed-forward neural network ($\operatorname{FFN}$). The output of each module is added to its input via a residual connection and fed into a normalization layer based on \cite{Ba_2016_arxiv},
\vspace{-0.2cm}
\begin{equation}
    \begin{gathered}
 \operatorname{Transformer_l}(I) = \operatorname{Norm}(I^{'} + \operatorname{FFN}(I^{'}))\\
 I^{'} = \operatorname{Norm}(I + \operatorname{Attn_{MH}}(I)).
\end{gathered}
\end{equation}
\vspace{-0.2cm}

\subsection{Semantic Attention Interaction Module (SAIM)}
\label{saim}

\begin{figure}[t]
\vspace{0.2cm}
\centering
\includegraphics[width=1\columnwidth]{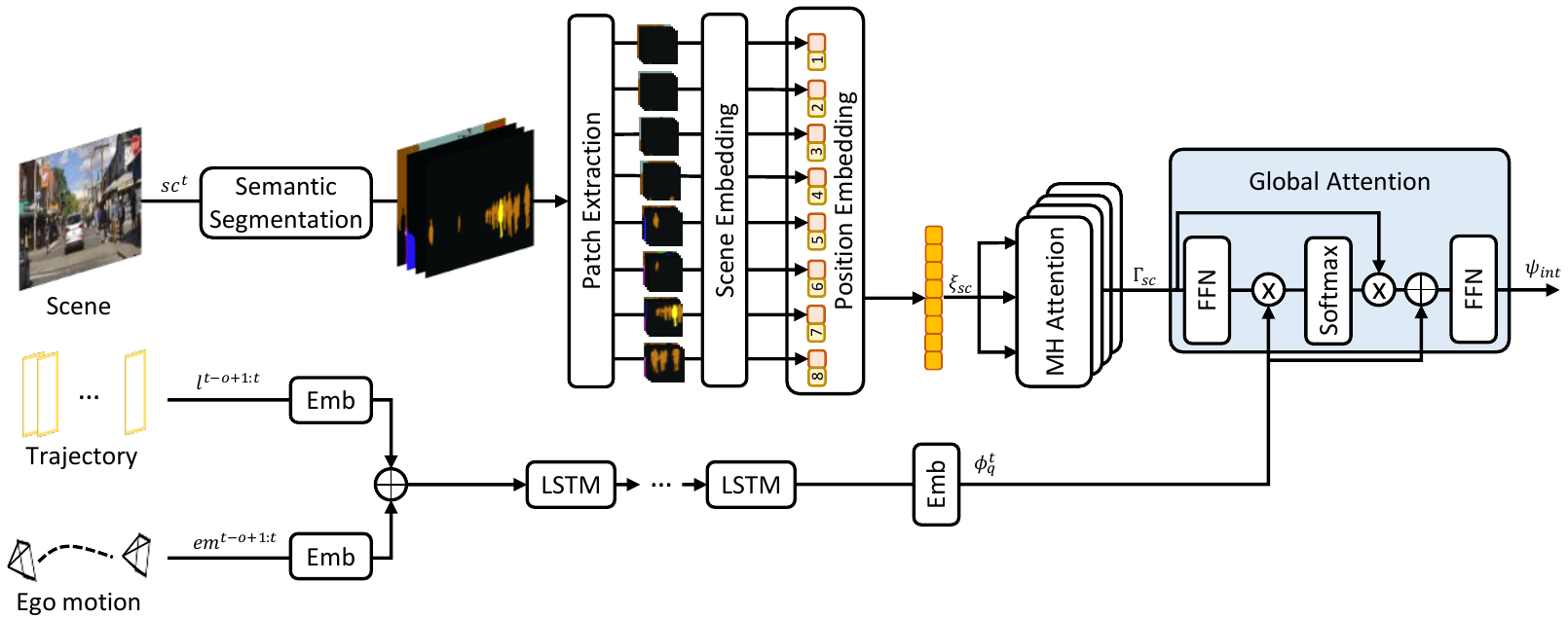}
\caption{An diagram of the semantic interaction attention module (SAIM) which takes as input scene image at time $t$, ego-vehicle motion and pedestrian trajectory. }
\vspace{-1.5em}
\label{saim}
\end{figure}

As demonstrated by the authors of \cite{Rasouli_2021_ICCV}, visual information can be used to effectively model interactions between pedestrians and their surroundings. Their approach, however, has a number of drawbacks. First, it requires sequences of global scenes grouped by semantic categories, each of which are processed separately using CNNs and recurrent networks. Second, sequences of pedestrians and surrounding people are encoded for each prediction which is computationally expensive. Third, ego-vehicle motion is not accounted for. We address these issues by using an attention-based architecture.

The input to the proposed SAIM module is the RGB image of current scene at time $t$ recorded from an ego-centric perspective. The image is passed to a segmentation algorithm whose output is organized into four channels: persons, bikes/bicyclists, vehicles and static context. The multi-channel map is then split into local patches of size $ps \times ps$ producing a total of $np$ patches. The patches are converted into vectors of size $\lambda_p$ via linear embedding layer. Inspired by \cite{Dosovitskiy_2021_ICLR}, the embedded patches are concatenated with positional encodings forming the input to an $\operatorname{Attn_{MH}}$ unit that outputs  $ \Gamma_{sc} \in R ^{np \times \lambda_p}$.  We encode pedestrian coordinates and ego-motion for the duration of observation using embedding layers and a shared LSTM unit. The output of the LSTM is embedded into vector $\phi^t_q$ and fed into a global attention module \cite{Luong_2015_arxiv} to generate the final interaction encoding $\psi_{int}$ as follows,
\vspace{-0.2cm}
\begin{equation}
    \begin{gathered}
\psi_{int} = tanh(W_c[c\oplus \phi^t_q])\\
c = softmax(\phi_q^tW_\Gamma \Gamma_{sc})\Gamma_{sc}\\
\Gamma_{sc} = \operatorname{Attn_{MH}}(\xi_{sc},\xi_{sc},\xi_{sc})
\end{gathered}
\vspace{-0.2cm}
\end{equation}

\noindent where $\xi_{sc}$ is the output of scene positional embedding.

\subsection{Hybrid Gated Decoder}
\label{decoder}
\textbf{Decoders.} In a multitask learning framework, capturing the correlation between decoders for individual tasks is essential. \cite{Rasouli_2021_ICCV} used two sets of decoders that independently and jointly produced predictions. However, this method, is computationally inefficient as two sets of predictions are generated and a post-processing stage is needed to combine them. Furthermore, there is no feedback between task-specific decoders and the shared one. 

Our proposed hybrid decoder architecture, motivated by \cite{Liu_2016_IJCAI}, also uses independent and shared decoders. Unlike \cite{Rasouli_2021_ICCV}, the shared decoder serves as a knowledge-base allowing the individual decoders to benefit from the shared information. To this end, the output of the shared decoder is passed through gate mechanism which controls the flow of information and is formulated as,
\vspace{-0.2cm}
\begin{equation}
    \begin{gathered}
 h_j^t = o_j^t \odot tanh(c_j^t)\\
 c_j^t = f_j^t \odot c^{t-1}_j  + i_j^t \odot \hat{c}^t_j \\
 \hat{c}_j^t = tanh(W_c\hat{x}^t) + U_ch^{t-1}_j) 
\end{gathered}
\vspace{-0.2cm}
\end{equation}

\noindent where $i^t_j$, $f^t_j$, and $o^t_j$ are input, forget and output gates of decoder $j$, $c^t_j$ and $h^t_j$ are its cell state and output at time $t$. Here, $h_{sd}^t = \overleftarrow{h_{sd}^t} \oplus \overrightarrow{h_{sd}^t} $ is the output of the shared decoder where $\overleftarrow{}$ and $\overrightarrow{}$ denote forward and backward LSTMs. The input $\hat{x}^t$ is given by,
\vspace{-0.2cm}

\begin{equation*}
    \begin{gathered}
  \hat{x}^{t} = \sigma(h_{sd}^t)\odot h_{sd}^t \oplus \Psi^t\\
  \Psi^t = \psi_{cm} \oplus \psi_{int} \oplus em^t. 
 \end{gathered}
 \vspace{-0.2cm}
\end{equation*}

\textbf{Prediction.} The primary tasks are trajectory and action prediction. Trajectory is predicted as 2D bounding box coordinates of the pedestrians by applying a linear transformation to the hidden states of the trajectory decoder. To predict the probability of crossing action at some time in the future, we perform a step-wise linear transformation at each time step followed by a sigmoid activation. For a unified prediction, we apply a global average pooling to compute the mean prediction across time steps. An auxiliary task, prediction of final discrete location of pedestrians, acts as a regularizer and a bridge between the two primary tasks. Similar to action, we classify the locations of the pedestrian at each time-step and average them over all prediction time steps.

\textbf{Learning Objectives.} Our learning objective is a weighted sum of the objectives for each task,
\begin{equation}
    \begin{gathered}\vspace{-0.3cm}
   L = \omega_l \sum_{i=1}^n \sum_{j=t}^{t+\tau} \log(\mathrm{cosh}(y^j_i - \hat{y}^j_i))\\ \vspace{-0.3cm}
    - \omega_a \sum_{i=1}^n y_i \log(\hat{y}_i) + (1-y_i)\log(1-\hat{y}_i) \\
    - \omega_{dl} \sum_{i=1}^i \sum_{dlc} y_{i,dlc} \log(\hat{y}_{i,dlc})
\end{gathered}
\end{equation}
\noindent where $\omega_l$, $\omega_a$ and $\omega_{dl}$ are weight coefficients for each task determined empirically, $dlc$ is the discrete location classes. 

\section{Evaluation}
\textbf{Implementation.} We set the sizes of all input and attention embeddings to $64$. For the encoder, we use $\operatorname{Attn_{MH}}$ and two transformer layers. In SAIM, semantic maps are generated using \cite{Chen_2017_arxiv} pre-trained on the CityScapes dataset \cite{Cordts_2016_CVPR} and four $\operatorname{Attn_{MH}}$. The maps are split into patches of size $12 \times 12$. The output dimension of the global attention is set to $128$.  

For LSTM units, we use $128$ hidden cells and L2 regularization of $0.0001$. We use $\mathrm{tanh}$ activation for SAIM and trajectory decoder and  $\mathrm{softsign}$ for remaining decoders. Shared decoder is implemented as a bidirectional LSTM. Following \cite{Rasouli_2021_ICCV}, we divided the image plane into a grid of $18 \times 32$ with $60 \times 60$ px cells for discrete locations.

\textbf{Datasets.} \textit{Pedestrian Intention Estimation (PIE)} \cite{Rasouli_2019_ICCV} is chosen as the primary dataset for evaluation as it contains over $6$ hours of ego-centric driving footage along with bounding box annotations for traffic objects, action labels for pedestrians and ego-vehicle sensor information. We use the default data split and sample pedestrians sequences with $50\%$ overlap and time to event between $1$ and $2$ seconds following \cite{Rasouli_2021_ICCV} resulting in $3980$ training sequences out of which $995$ are crossing events.

\textit{Joint Attention in Autonomous Driving (JAAD)} \cite{Rasouli_2017_ICCVW} is a pedestrian dataset with short ego-centric driving clips of urban driving with bounding box and action annotations for pedestrians. Unlike the PIE dataset, JAAD lacks ego-motion information, therefore we use textual driver action labels, such as \textit{moving slow} and \textit{slowing down}, instead. We use the data split proposed in \cite{Rasouli_2018_ECCVW} and sample sequences similar to PIE resulting in $3955$ sequences with $807$ crossing samples.

\textbf{Training.} The proposed model is trained with the initial learning rate of $10^{-4}$ on PIE and $5\times 10^{-5}$ on JAAD and batch size of $8$ using RMSProp \cite{Tieleman_2012_tech} for $200$ epochs. Learning rate is reduced by a factor of $0.2$ based on the validation loss. To compensate for crossing/non-crossing sample imbalance, we used class weights computed using the ratio of the samples. For loss weights we set the following values empirically, $\omega_l=0.6$ for PIE and $\omega_l=0.5$ for JAAD with $\omega_a=1$ and $\omega_g=1$ same for both datasets. 

\textbf{Metrics.} We report the results for $0.5$s observation and $1$s prediction. For \textit{trajectory prediction}, we use $\mathbf{ADE}$ and $\mathbf{FDE}$ metrics computed based on the center coordinates of pedestrian bounding boxes, and  $\mathbf{ARB}$ and $\mathbf{FRB}$ calculated by RMSE of bounding box coordinates. The latter shows how well the model localizes pedestrians in image plane. All metrics are reported in pixels. For \textit{action prediction}, we use common classification metrics including $\mathbf{accuracy}$, Area Under Curve ($\mathbf{AUC}$), $\mathbf{F1}$ and $\mathbf{precision}$.

\textbf{Models.} We compare our model to state-of-the-art ego-centric behavior prediction methods. For \textit{trajectory prediction} we use the following methods: \textit{Future Person Localization (FPL)} (only predicts center coordinates) \cite{Yagi_2018_CVPR}, 
\textit{Bayesian LSTM (B-LSTM)} \cite{Bhattacharyya_2018_CVPR}, \textit{FOL} \cite{Yao_2019_ICRA}, and \textit{PIE$_{traj}$} and \textit{PIE$_{full}$} introduced in \cite{Rasouli_2019_ICCV}. For \textit{action prediction} the following models are selected: \textit{ATGC} \cite{Rasouli_2017_ICCVW}, \textit{MM-LSTM} \cite{Aliakbarian_2018_ACCV}, \textit{SF-GRU} \cite{Rasouli_2019_BMVC}, \textit{PCPA} \cite{Kotseruba_2021_WACV}, and \textit{I3D} \cite{Carreira_2017_CVPR}. We also compare to state-of-the-art \textit{multitask prediction} model, \textit{BiPed} \cite{Rasouli_2021_ICCV}.

\vspace{-0.2cm}
\subsection{Behavior Prediction}
\label{pie_experiment}
\begin{table}[!t]
\vspace{+0.1cm}
\caption{Comparison to SOTA methods. $\uparrow$ and $\downarrow$ mean higher or lower values are better respectively.}

\label{sota_results}
\centering
\resizebox{1\columnwidth}{!}{%
\begin{tabular}{lcccc?lcccc}
\multicolumn{5}{c?}{Trajectory}                                  & \multicolumn{5}{c}{Action}                               \\ \hline
\multicolumn{1}{l||}{Method}    & ADE   & FDE    & ARB   & FRB    & \multicolumn{1}{l||}{Method}  & Acc  & AUC  & F1   & Prec \\ \hline
\multicolumn{10}{c}{\textbf{PIE}}   \\ \hline
\multicolumn{1}{l||}{FOL}       & 73.87 & 164.53 & 78.16 & 143.69 & \multicolumn{1}{l||}{ATGC}    & 0.59 & 0.55 & 0.36 & 0.35 \\
\multicolumn{1}{l||}{FPL}       & 56.66 & 132.23 & -     & -      & \multicolumn{1}{l||}{I3D}     & 0.79 & 0.75 & 0.64 & 0.61 \\
\multicolumn{1}{l||}{B-LSTM}    & 27.09 & 66.74  & 37.41 & 75.87  & \multicolumn{1}{l||}{MM-LSTM} & 0.84 & 0.84 & 0.75 & 0.68 \\
\multicolumn{1}{l||}{PIE$_{traj}$}   & 21.82 & 53.63  & 27.16 & 55.39  & \multicolumn{1}{l||}{SF-GRU}  & 0.86 & 0.83 & 0.75 & 0.73 \\
\multicolumn{1}{l||}{PIE$_{full}$}   & 19.50 & 45.27  & 24.40 & 49.09  & \multicolumn{1}{l||}{PCPA}    & 0.86 & 0.84 & 0.76 & 0.73 \\ \hline
\multicolumn{1}{l||}{BiPed}     & 15.21 & 35.03  & 19.62 & 39.12  & \multicolumn{1}{l||}{BiPed}        & 0.91 & \textbf{0.90} & 0.85 & 0.82 \\ \hline
\multicolumn{1}{l||}{\textbf{PedFormer}} &    \textbf{13.08}   & \textbf{30.35}       & \textbf{15.27}      & \textbf{32.79}       & \multicolumn{1}{l||}{\textbf{PedFormer}}        & \textbf{0.93}     &  \textbf{0.90}    & \textbf{0.87}      & \textbf{0.89} \\ \hline
\multicolumn{10}{c}{\textbf{JAAD}}   \\ \hline
\multicolumn{1}{l||}{FOL}       & 61.39 & 126.97 & 70.12 & 129.17 & \multicolumn{1}{l||}{ATGC}    & 0.64 & 0.60 & 0.53 & 0.50 \\
\multicolumn{1}{l||}{FPL}       & 42.24 & 86.13 & -     & -      & \multicolumn{1}{l||}{I3D}      & 0.82 & 0.75 & 0.55 & 0.49 \\
\multicolumn{1}{l||}{B-LSTM}    & 28.36 & 70.22  & 39.14 & 79.66  & \multicolumn{1}{l||}{MM-LSTM} & 0.80 & 0.60 & 0.40 & 0.39 \\
\multicolumn{1}{l||}{PIE$_{traj}$}   & 23.49 & 50.18  & 30.40 & 57.17  & \multicolumn{1}{l||}{SF-GRU}  & 0.83 & 0.77 & 0.58 & 0.51 \\
\multicolumn{1}{l||}{PIE$_{full}$}   & 22.83 & 49.44  & 29.52 & 55.43  & \multicolumn{1}{l||}{PCPA}    & 0.83 & 0.77 & 0.57 & 0.50 \\ \hline
\multicolumn{1}{l||}{BiPed}     & 20.58 & 46.85  & 27.98 & 55.07  & \multicolumn{1}{l||}{BiPed}        & 0.83 & \textbf{0.79} & \textbf{0.60} & 0.52 \\ \hline 
\multicolumn{1}{l||}{\textbf{PedFormer}} &  \textbf{17.89}&   \textbf{41.63}  & \textbf{24.56}   & \textbf{48.82}        & \multicolumn{1}{l||}{\textbf{PedFormer}} &    \textbf{0.93}    &0.76        & 0.54      & \textbf{0.65} 
\end{tabular}}
\vspace{-0.4cm}
\end{table}

\textbf{PIE}. According to Table \ref{sota_results}, our model achieves state-of-the-art performance on all metrics. Trajectory prediction is improved by $22\%$ on localization (bounding box) metrics and $14\%$ on the rest. The most notable improvement on action prediction is on $\mathit{precision}$ where a performance boost of $7\%$ is gained. Such a boost highlights the effectiveness of our method in capturing the dependency between complementary tasks, in this case using predicted trajectory to distinguish between crossing and non-crossing actions.

\textbf{JAAD}. On JAAD dataset, significant improvement of up to $13\%$ is achieved across different metrics for both trajectory and action prediction. Compared to PIE, improvement on localization metrics is more limited primarily due to the homogeneity of JAAD in which the majority of samples are the pedestrians that are walking on sidewalks towards or away from the camera. On actions,  the biggest improvement is achieved on $\mathit{precision}$ (as is the case for PIE dataset), while $\mathit{accuracy}$ is also improved by up to $10\%$.     

\vspace{-0.2cm}
\subsection{Ablation studies}
\label{ablation_study}
We report the results on the core behavior task of trajectory prediction and elaborate on actions as part of the discussion. We also use a more explicit localization metric, final intersection over union ($\mathbf{FIoU}$), to highlight how well the final predicted bounding boxes overlap with the ground truth. 

\begin{table}[!t]
\vspace{0.1cm}
\centering
\caption{Performance of different encoding methods. $\uparrow$ and $\downarrow$ mean higher or lower values are better respectively. }
\label{ablation_encoding}
\resizebox{1\columnwidth}{!}{%
\begin{tabular}{l||cc|ccccc}
Methods                   & \multicolumn{2}{c|}{Interaction} & $ADE\downarrow$  &$FDE\downarrow$  & $ARB\downarrow$ & $FRB\downarrow$ & $FIoU\uparrow$ \\ \hline
Modality  Transformers       			 &
                        \multicolumn{2}{c|}{\xmark}&
                 15.07  &  36.28  & 20.16   &  40.91  & 0.103   \\       
Shared  Transformer    			 &
                        \multicolumn{2}{c|}{\xmark}&

              14.88   &  35.79  & 19.91	&  40.03  &0.108  \\
PedFormer       &
            	            \multicolumn{2}{c|}{\xmark}&
13.60	&  32.06	  & 17.28	&  35.27  & 0.111 \\ \hline
PedFormer w/o G Att&
                         \multicolumn{2}{c|}{\cmark}&
14.14	& 33.57   &  18.95   &   38.06 &    0.112 \\ 

PedFormer w/o Motion&
\multicolumn{2}{c|}{\cmark}&
             13.51    & 31.48   &  17.31   &   35.22 &   0.117  \\ 

\textbf{PedFormer}  			 &
            \multicolumn{2}{c|}{\cmark}&
\textbf{13.08}   & \textbf{30.35}       & \textbf{15.27}      & \textbf{32.79}  & \textbf{0.122}      \\ 
\end{tabular}
}
\vspace{-0.4cm}
\end{table}

\begin{figure*}[ht!]
\vspace{0.2cm}
\includegraphics[width=1\textwidth]{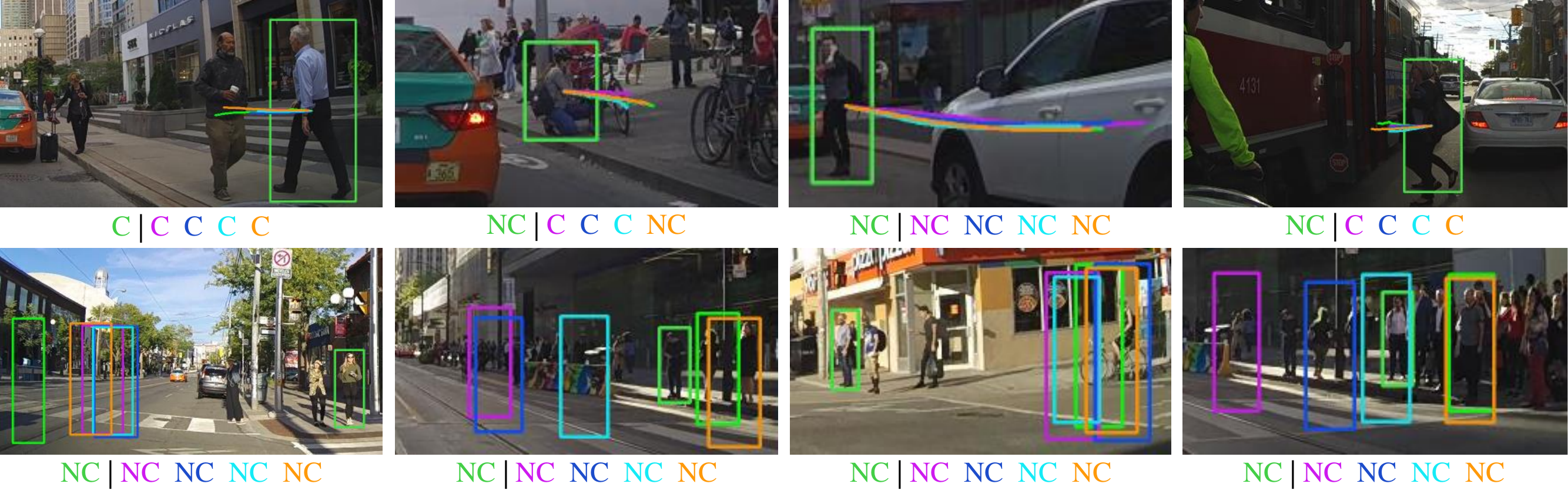}
\caption{Qualitative results of PedFormer on PIE. First row shows predicted trajectories and second final predicted bounding boxes. The colors correspond to {\color{gt}ground truth}, {\color{pedformer} PedFormer (\textbf{ours})}, {\color{multi_trans} Modality Trans.}, {\color{sep_dec} Task-based} decoder and PedFormer with {\color{no_int} no interaction}. The letters indicate predicted crossing (C) or non-crossing (NC) events.}
\label{qualitative}
\vspace{-0.4cm}
\end{figure*}

\textbf{Multimomdal encoding.} We report on alternative transformer-based encoding schemes, namely \textit{modality transformer} where each modality is passed through independent transformer and then combined at the end, and \textit{shared transformer} in which the input modalities are first combined before entering a shared transformer for encoding. For these variations no interaction module is used and the rest of the model remains the same. 

The results of the experiment are shown in Table \ref{ablation_encoding}. First observation is that, overall, using separate encoders is inferior as the model fails to capture the correlation between different modalities. Simply sharing the encoder, however, does not result in significant improvements. Once pair-wise cross-modal attention scoring is added, a significant boost across all metrics can be observed. The greater improvement over $\mathit{ARB/FRB}$ indicates that the model better captures the relation between the scale of bounding boxes and ego-motion which directly affects it.       

\textbf{Interaction modeling.} As a baseline, we omit the global attention component and replace it with a $1 \times 1$ convolutional layer to generate a summary vector over the output of the multi-head attention modules denoted as \textit{PedFormer w/o G. Att}. To highlight the importance of dynamics in interaction modeling, we also remove the motion encoding (\textit{PedFormer w/o Motion}) and use the last element of the multi-head attention output as a reference to generate a summary.

As shown in Table \ref{ablation_encoding}, simply using an attention framework over scenes not only does not provide any benefits but can also add noise and consequently deteriorate the results. Having a dual attention framework mitigates the issue and results in measurable, albeit limited, improvement. This is due to the fact that although a single image provides information regarding the structure of the scene, on its own it does not capture the movements of the agents. Consequently, once dynamic information is included in the computation of the final representation, a significant improvement is achieved.  Overall, using the proposed interaction modeling technique provides up to $12\%$ improvement over trajectory metrics and $3\%$ on action metrics, specifically $\mathit{AUC}$ and $\mathit{F1}$.

\begin{table}[t]
\caption{Performance of different decoding mechanisms. $\uparrow$ and $\downarrow$ mean higher or lower values are better respectively.}
\label{ablation_decoding}
\centering
\resizebox{1\columnwidth}{!}{%
\begin{tabular}{l||ccccc}
            Method                           & $ADE\downarrow$  &$FDE\downarrow$  & $ARB\downarrow$ & $FRB\downarrow$ & $FIoU\uparrow$\\ \hline
Task-based     &  16.10	&   37.79&  21.22&	42.76 & 0.100\\
Shared	 	    &  15.12&	35.44&	19.49&	39.33  & 0.100\\
Bifold          &  14.83&	35.23&	18.93&	38.04  & 0.105 \\
Hybrid   	&  13.78&	32.89&	17.82&   36.51  & 0.110\\
\textbf{Gated Hybrid }  & \textbf{13.08}   & \textbf{30.35}       & \textbf{15.27}      & \textbf{32.79}  & \textbf{0.122} \\ 
\end{tabular}%
}
\vspace{-0.5cm}
\end{table}

\textbf{Decoding. } To study the impact of different decoding techniques on the overall performance of the model we use four variants of decoder architectures while keeping the rest of the model unchanged. The baseline models are \textit{task-based} where each task has a separate decoder and \textit{shared} where a single decoder is used for all tasks. We also use the \textbf{bifold} method of \cite{Rasouli_2021_ICCV} and two variations of the proposed \textit{hybrid} method with and without a gating mechanism. 

Results in Table \ref{ablation_decoding} show that using a combined representation, whether bifold or hybrid, leads to better performance on all trajectory and action metrics with the biggest improvement achieved on $\mathit{precision}$, which agrees with findings in \cite{Rasouli_2021_ICCV}. Direct feedback mechanism between shared and independent decoders allows hybrid approach to better learn the dependencies between tasks compared to bifold method. Since uncontrolled flow of information into task decoders can introduce noise,  proposed gating mechanism allows the task decoders to be selective on the use of shared representation and provides further improvements. Overall, compared to the baseline task-based method, the proposed gated hybrid approach improves the performance by up to $12\%$ on trajectory and $4\%$ on action metrics.

\vspace{-0.2cm}
\subsection{Qualitative results}
Figure \ref{qualitative} shows qualitative examples of the proposed and baseline methods. PedFormer consistently produces better predictions, both in terms of trajectories and bounding boxes, in various scenarios. It is particularly evident when ego-motion is significant, e.g. when the ego-vehicle is making a right turn or is accelerating (first and second images from the left in bottom row), owing to better encoding of ego-motion and effective use of visual context. In the case of crossing events, sometimes the trajectory can be misleading, e.g. the top right image shows the pedestrian boarding a streetcar but all models classified the event as crossing due to the presence of lateral movement towards the road.

\section{Conclusion}
This paper introduced a novel pedestrian behavior prediction method that simultaneously predicts trajectories and actions. The proposed approach benefits from a cross-modal transformer-based encoder, a scene-based interaction module with attention modulation, and a gated hybrid decoder mechanism for producing predictions. We conducted experiments on public benchmark datasets and showed that our method achieves state-of-the-art performance on both tasks by a wide margin. Furthermore, via ablation studies, we highlighted the contribution of different proposed mechanisms on the overall performance of our method.

\bibliographystyle{IEEEtran}
\bibliography{refs}

\begin{thebibliography}{10}
\providecommand{\url}[1]{#1}
\csname url@rmstyle\endcsname
\providecommand{\newblock}{\relax}
\providecommand{\bibinfo}[2]{#2}
\providecommand\BIBentrySTDinterwordspacing{\spaceskip=0pt\relax}
\providecommand\BIBentryALTinterwordstretchfactor{4}
\providecommand\BIBentryALTinterwordspacing{\spaceskip=\fontdimen2\font plus
\BIBentryALTinterwordstretchfactor\fontdimen3\font minus
  \fontdimen4\font\relax}
\providecommand\BIBforeignlanguage[2]{{%
\expandafter\ifx\csname l@#1\endcsname\relax
\typeout{** WARNING: IEEEtran.bst: No hyphenation pattern has been}%
\typeout{** loaded for the language `#1'. Using the pattern for}%
\typeout{** the default language instead.}%
\else
\language=\csname l@#1\endcsname
\fi
#2}}

\bibitem{Rasouli_2019_ITS}
A.~Rasouli and J.~K. Tsotsos, ``Autonomous vehicles that interact with
  pedestrians: A survey of theory and practice,'' \emph{IEEE Transactions on
  Intelligent Transportation Systems}, vol.~21, no.~3, pp. 900--918, 2019.

\bibitem{Rasouli_2017_IV}
A.~Rasouli, I.~Kotseruba, and J.~K. Tsotsos, ``Agreeing to cross: How drivers
  and pedestrians communicate,'' in \emph{Intelligent Vehicles Symposium (IV)},
  2017.

\bibitem{Shi_2021_CVPR}
L.~Shi, L.~Wang, C.~Long, S.~Zhou, M.~Zhou, Z.~Niu, and G.~Hua, ``Sgcn: Sparse
  graph convolution network for pedestrian trajectory prediction,'' in
  \emph{CVPR}, 2021.

\bibitem{Rasouli_2019_ICCV}
A.~Rasouli, I.~Kotseruba, T.~Kunic, and J.~K. Tsotsos, ``{PIE}: A large-scale
  dataset and models for pedestrian intention estimation and trajectory
  prediction,'' in \emph{ICCV}, 2019.

\bibitem{Liu_2020_RAL}
B.~Liu, E.~Adeli, Z.~Cao, K.-H. Lee, A.~Shenoi, A.~Gaidon, and J.~C. Niebles,
  ``Spatiotemporal relationship reasoning for pedestrian intent prediction,''
  \emph{IEEE Robotics and Automation Letters}, vol.~5, no.~2, pp. 3485--3492,
  2020.

\bibitem{Rasouli_2019_BMVC}
A.~Rasouli, I.~Kotseruba, and J.~K. Tsotsos, ``Pedestrian action anticipation
  using contextual feature fusion in stacked {RNN}s,'' in \emph{BMVC}, 2019.

\bibitem{Rasouli_2021_ICCV}
A.~Rasouli, M.~Rohani, and J.~Luo, ``Bifold and semantic reasoning for
  pedestrian behavior prediction,'' in \emph{ICCV}, 2021.

\bibitem{Liang_2019_CVPR}
J.~Liang, L.~Jiang, J.~C. Niebles, A.~G. Hauptmann, and L.~Fei-Fei, ``Peeking
  into the future: Predicting future person activities and locations in
  videos,'' in \emph{CVPR}, 2019.

\bibitem{Rasouli_2017_ICCVW}
A.~Rasouli, I.~Kotseruba, and J.~K. Tsotsos, ``Are they going to cross? {A}
  benchmark dataset and baseline for pedestrian crosswalk behavior,'' in
  \emph{ICCVW}, 2017.

\bibitem{Liu_2020_ECCV}
M.~Liu, S.~Tang, Y.~Li, and J.~Rehg, ``Forecasting human object interaction:
  Joint prediction of motor attention and actions in first person video,'' in
  \emph{ECCV}, 2020.

\bibitem{Piergiovanni_2020_ECCV}
A.~Piergiovanni, A.~Angelova, A.~Toshev, and M.~S. Ryoo, ``Adversarial
  generative grammars for human activity prediction,'' in \emph{ECCV}, 2020.

\bibitem{Joo_2019_CVPR}
H.~Joo, T.~Simon, M.~Cikara, and Y.~Sheikh, ``Towards social artificial
  intelligence: Nonverbal social signal prediction in a triadic interaction,''
  in \emph{CVPR}, 2019.

\bibitem{Yao_2018_CVPR}
T.~Yao, M.~Wang, B.~Ni, H.~Wei, and X.~Yang, ``Multiple granularity group
  interaction prediction,'' in \emph{CVPR}, 2018.

\bibitem{Epstein_2020_CVPR}
D.~Epstein, B.~Chen, and C.~Vondrick, ``Oops! {P}redicting unintentional action
  in video,'' in \emph{CVPR}, 2020.

\bibitem{Qi_2020_CVPR}
M.~Qi, J.~Qin, Y.~Wu, and Y.~Yang, ``Imitative non-autoregressive modeling for
  trajectory forecasting and imputation,'' in \emph{CVPR}, 2020.

\bibitem{Felsen_2017_ICCV}
P.~Felsen, P.~Agrawal, and J.~Malik, ``What will happen next? {F}orecasting
  player moves in sports videos,'' in \emph{ICCV}, 2017.

\bibitem{Dendorfer_2021_ICCV}
P.~Dendorfer, S.~Elflein, and L.~Leal-Taixe, ``Mg-gan: A multi-generator model
  preventing out-of-distribution samples in pedestrian trajectory prediction,''
  in \emph{ICCV}, 2021.

\bibitem{Shafiee_2021_CVPR}
N.~Shafiee, T.~Padir, and E.~Elhamifar, ``Introvert: Human trajectory
  prediction via conditional 3d attention,'' in \emph{CVPR}, 2021.

\bibitem{Hu_2020_CVPR}
Y.~Hu, S.~Chen, Y.~Zhang, and X.~Gu, ``Collaborative motion prediction via
  neural motion message passing,'' in \emph{CVPR}, 2020.

\bibitem{Mohamed_2020_CVPR}
A.~Mohamed, K.~Qian, M.~Elhoseiny, and C.~Claudel, ``{Social-STGCNN}: A social
  spatio-temporal graph convolutional neural network for human trajectory
  prediction,'' in \emph{CVPR}, 2020.

\bibitem{Sun_2020_CVPR}
J.~Sun, Q.~Jiang, and C.~Lu, ``Recursive social behavior graph for trajectory
  prediction,'' in \emph{CVPR}, 2020.

\bibitem{Sun_2020_CVPR_2}
H.~Sun, Z.~Zhao, and Z.~He, ``Reciprocal learning networks for human trajectory
  prediction,'' in \emph{CVPR}, 2020.

\bibitem{Mangalam_2020_ECCV}
K.~Mangalam, H.~Girase, S.~Agarwal, K.-H. Lee, E.~Adeli, J.~Malik, and
  A.~Gaidon, ``It is not the journey but the destination: Endpoint conditioned
  trajectory prediction,'' in \emph{ECCV}, 2020.

\bibitem{Choi_2019_ICCV}
C.~Choi and B.~Dariush, ``Looking to relations for future trajectory
  forecast,'' in \emph{ICCV}, 2019.

\bibitem{Zhang_2019_CVPR}
P.~Zhang, W.~Ouyang, P.~Zhang, J.~Xue, and N.~Zheng, ``{SR-LSTM}: State
  refinement for {LSTM} towards pedestrian trajectory prediction,'' in
  \emph{CVPR}, 2019.

\bibitem{Sadeghian_2019_CVPR}
A.~Sadeghian, V.~Kosaraju, A.~Sadeghian, N.~Hirose, H.~Rezatofighi, and
  S.~Savarese, ``{SoPhie}: An attentive {GAN} for predicting paths compliant to
  social and physical constraints,'' in \emph{CVPR}, 2019.

\bibitem{Gupta_2018_CVPR}
A.~Gupta, J.~Johnson, L.~Fei-Fei, S.~Savarese, and A.~Alahi, ``Social {GAN}:
  Socially acceptable trajectories with generative adversarial networks,'' in
  \emph{CVPR}, 2018.

\bibitem{Neumann_2021_CVPR}
L.~Neumann and A.~Vedaldi, ``Pedestrian and ego-vehicle trajectory prediction
  from monocular camera,'' in \emph{CVPR}, 2021.

\bibitem{Makansi_2020_CVPR}
O.~Makansi, O.~Cicek, K.~Buchicchio, and T.~Brox, ``Multimodal future
  localization and emergence prediction for objects in egocentric view with a
  reachability prior,'' in \emph{CVPR}, 2020.

\bibitem{Malla_2020_CVPR}
S.~Malla, B.~Dariush, and C.~Choi, ``{TITAN}: Future forecast using action
  priors,'' in \emph{CVPR}, 2020.

\bibitem{Yagi_2018_CVPR}
T.~Yagi, K.~Mangalam, R.~Yonetani, and Y.~Sato, ``Future person localization in
  first-person videos,'' in \emph{CVPR}, 2018.

\bibitem{Yao_2019_ICRA}
Y.~Yao, M.~Xu, C.~Choi, D.~J. Crandall, E.~M. Atkins, and B.~Dariush,
  ``Egocentric vision-based future vehicle localization for intelligent driving
  assistance systems,'' in \emph{ICRA}, 2019.

\bibitem{Yao_2019_IROS}
Y.~Yao, M.~Xu, Y.~Wang, D.~J. Crandall, and E.~M. Atkins, ``Unsupervised
  traffic accident detection in first-person videos,'' in \emph{IROS}, 2019.

\bibitem{Bhattacharyya_2018_CVPR}
A.~Bhattacharyya, M.~Fritz, and B.~Schiele, ``Long-term on-board prediction of
  people in traffic scenes under uncertainty,'' in \emph{CVPR}, 2018.

\bibitem{Chandra_2019_CVPR}
R.~Chandra, U.~Bhattacharya, A.~Bera, and D.~Manocha, ``{TraPHic}: Trajectory
  prediction in dense and heterogeneous traffic using weighted interactions,''
  in \emph{CVPR}, 2019.

\bibitem{Kotseruba_2021_WACV}
I.~Kotseruba, A.~Rasouli, and J.~K. Tsotsos, ``Benchmark for evaluating
  pedestrian action prediction,'' in \emph{WACV}, 2021.

\bibitem{Chaabane_2020_WACV}
M.~Chaabane, A.~Trabelsi, N.~Blanchard, and R.~Beveridge, ``Looking ahead:
  Anticipating pedestrians crossing with future frames prediction,'' in
  \emph{WACV}, 2020.

\bibitem{Saleh_2019_ICRA}
K.~Saleh, M.~Hossny, and S.~Nahavandi, ``Real-time intent prediction of
  pedestrians for autonomous ground vehicles via spatio-temporal {DenseNet},''
  in \emph{ICRA}, 2019.

\bibitem{Gujjar_2019_ICRA}
P.~Gujjar and R.~Vaughan, ``Classifying pedestrian actions in advance using
  predicted video of urban driving scenes,'' in \emph{ICRA}, 2019.

\bibitem{Aliakbarian_2018_ACCV}
M.~S. Aliakbarian, F.~S. Saleh, M.~Salzmann, B.~Fernando, L.~Petersson, and
  L.~Andersson, ``{VIENA}: A driving anticipation dataset,'' in \emph{ACCV},
  2019.

\bibitem{Luvizon_2018_CVPR}
D.~C. Luvizon, D.~Picard, and H.~Tabia, ``{2D/3D} pose estimation and action
  recognition using multitask deep learning,'' in \emph{CVPR}, 2018.

\bibitem{Guo_2018_ECCV}
M.~Guo, A.~Haque, D.-A. Huang, S.~Yeung, and L.~Fei-Fei, ``Dynamic task
  prioritization for multitask learning,'' in \emph{ECCV}, 2018.

\bibitem{Hu_2018_ECCV}
G.~Hu, L.~Liu, Y.~Yuan, Z.~Yu, Y.~Hua, Z.~Zhang, F.~Shen, L.~Shao,
  T.~Hospedales, N.~Robertson, and Y.~Yang, ``Deep multi-task learning to
  recognise subtle facial expressions of mental states,'' in \emph{ECCV}, 2018.

\bibitem{Du_2019_CVPR}
K.~Du, X.~Lin, Y.~Sun, and X.~Ma, ``{CrossInfoNet}: Multi-task information
  sharing based hand pose estimation,'' in \emph{CVPR}, 2019.

\bibitem{Mallya_2018_ECCV}
A.~Mallya, D.~Davis, and S.~Lazebnik, ``Piggyback: Adapting a single network to
  multiple tasks by learning to mask weights,'' in \emph{ECCV}, 2018.

\bibitem{Zeng_2019_ICCV}
Z.~Zeng, X.~Li, Y.~K. Yu, and C.-W. Fu, ``Deep floor plan recognition using a
  multi-task network with room-boundary-guided attention,'' in \emph{ICCV},
  2019.

\bibitem{Tang_2019_ICCV}
Z.~Tang, M.~Naphade, S.~Birchfield, J.~Tremblay, W.~Hodge, R.~Kumar, S.~Wang,
  and X.~Yang, ``{PAMTRI}: Pose-aware multi-task learning for vehicle
  re-identification using highly randomized synthetic data,'' in \emph{ICCV},
  2019.

\bibitem{Hassani_2019_ICCV}
K.~Hassani and M.~Haley, ``Unsupervised multi-task feature learning on point
  clouds,'' in \emph{ICCV}, 2019.

\bibitem{Casas_2018_CORL}
S.~Casas, W.~Luo, and R.~Urtasun, ``{IntentNet}: Learning to predict intention
  from raw sensor data,'' in \emph{CORL}, 2018.

\bibitem{Kendall_2018_CVPR}
A.~Kendall, Y.~Gal, and R.~Cipolla, ``Multi-task learning using uncertainty to
  weigh losses for scene geometry and semantics,'' in \emph{CVPR}, 2018.

\bibitem{Liang_2019_CVPR_2}
M.~Liang, B.~Yang, Y.~Chen, R.~Hu, and R.~Urtasun, ``Multi-task multi-sensor
  fusion for {3D} object detection,'' in \emph{CVPR}, 2019.

\bibitem{Liu_2019_CVPR}
S.~Liu, E.~Johns, and A.~J. Davison, ``End-to-end multi-task learning with
  attention,'' in \emph{CVPR}, 2019.

\bibitem{Wu_2020_CVPR_2}
P.~Wu, S.~Chen, and D.~N. Metaxas, ``{MotionNet}: Joint perception and motion
  prediction for autonomous driving based on bird's eye view maps,'' in
  \emph{CVPR}, 2020.

\bibitem{Hasan_2018_CVPR}
I.~Hasan, F.~Setti, T.~Tsesmelis, A.~Del~Bue, F.~Galasso, and M.~Cristani,
  ``{MX-LSTM}: Mixing tracklets and vislets to jointly forecast trajectories
  and head poses,'' in \emph{CVPR}, 2018.

\bibitem{Fernando_2018_ACCV}
T.~Fernando, S.~Denman, S.~Sridharan, and C.~Fookes, ``{GD-GAN}: Generative
  adversarial networks for trajectory prediction and group detection in
  crowds,'' in \emph{ACCV}, 2019.

\bibitem{Zhang_2020_CVPR}
Z.~Zhang, J.~Gao, J.~Mao, Y.~Liu, D.~Anguelov, and C.~Li, ``{STINet}:
  Spatio-temporal-interactive network for pedestrian detection and trajectory
  prediction,'' in \emph{CVPR}, 2020.

\bibitem{Choi_2021_CVPR}
C.~Choi, J.~H. Choi, J.~Li, and S.~Malla, ``Shared cross-modal trajectory
  prediction for autonomous driving,'' in \emph{CVPR}, 2021.

\bibitem{Park_2020_ECCV}
S.~H. Park, G.~Lee, M.~Bhat, J.~Seo, M.~Kang, J.~Francis, A.~R. Jadhav, P.~P.
  Liang, and L.-P. Morency, ``Diverse and admissible trajectory forecasting
  through multimodal context understanding,'' in \emph{ECCV}, 2020.

\bibitem{Yau_2021_ICRA}
T.~Yau, S.~Malekmohammadi, A.~Rasouli, P.~Lakner, M.~Rohani, and J.~Luo,
  ``Graph-sim: A graph-based spatiotemporal interaction modelling for
  pedestrian action prediction,'' in \emph{ICRA}, 2021.

\bibitem{Li_2020_NeurIPS}
J.~Li, F.~Yang, M.~Tomizuka, and C.~Choi, ``{EvolveGraph}: Multi-agent
  trajectory prediction with dynamic relational reasoning,'' in \emph{NeurIPS},
  2020.

\bibitem{Kosaraju_2019_NeurIPS}
V.~Kosaraju, A.~Sadeghian, R.~Martin-Martin, I.~Reid, H.~Rezatofighi, and
  S.~Savarese, ``{Social-BIGAT}: Multimodal trajectory forecasting using
  {B}icycle-{GAN} and graph attention networks,'' in \emph{NeurIPS}, 2019.

\bibitem{Kipf_2018_ICML}
T.~Kipf, E.~Fetaya, K.-C. Wang, M.~Welling, and R.~Zemel, ``Neural relational
  inference for interacting systems,'' in \emph{ICML}, 2018.

\bibitem{Ruder_2017_arxiv}
S.~Ruder, ``An overview of multi-task learning in deep neural networks,''
  \emph{arXiv:1706.05098}, 2017.

\bibitem{Vaswani_2017_NIPS}
A.~Vaswani, N.~Shazeer, N.~Parmar, J.~Uszkoreit, L.~Jones, A.~N. Gomez,
  {\L}.~Kaiser, and I.~Polosukhin, ``Attention is all you need,'' in
  \emph{NeurIPS}, 2017.

\bibitem{Ba_2016_arxiv}
J.~L. Ba, J.~R. Kiros, and G.~E. Hinton, ``Layer normalization,''
  \emph{arXiv:1607.06450}, 2016.

\bibitem{Dosovitskiy_2021_ICLR}
A.~Dosovitskiy, L.~Beyer, A.~Kolesnikov, D.~Weissenborn, X.~Zhai,
  T.~Unterthiner, M.~Dehghani, M.~Minderer, G.~Heigold, S.~Gelly,
  \emph{et~al.}, ``An image is worth 16x16 words: Transformers for image
  recognition at scale,'' in \emph{ICLR}, 2021.

\bibitem{Luong_2015_arxiv}
M.-T. Luong, H.~Pham, and C.~D. Manning, ``Effective approaches to
  attention-based neural machine translation,'' \emph{arXiv:1508.04025}, 2015.

\bibitem{Liu_2016_IJCAI}
P.~Liu, X.~Qiu, and X.~Huang, ``Recurrent neural network for text
  classification with multi-task learning,'' in \emph{IJCAI}, 2016.

\bibitem{Chen_2017_arxiv}
L.-C. Chen, G.~Papandreou, F.~Schroff, and H.~Adam, ``Rethinking atrous
  convolution for semantic image segmentation,'' \emph{arXiv:1706.05587}, 2017.

\bibitem{Cordts_2016_CVPR}
M.~Cordts, M.~Omran, S.~Ramos, T.~Rehfeld, M.~Enzweiler, R.~Benenson,
  U.~Franke, S.~Roth, and B.~Schiele, ``The {CityScapes} dataset for semantic
  urban scene understanding,'' in \emph{CVPR}, 2016.

\bibitem{Rasouli_2018_ECCVW}
A.~Rasouli, I.~Kotseruba, and J.~K. Tsotsos, ``It's not all about size: On the
  role of data properties in pedestrian detection,'' in \emph{ECCVW}, 2018.

\bibitem{Tieleman_2012_tech}
T.~Tieleman and G.~Hinton, ``Lecture 6.5-{RMSProp}, coursera: Neural networks
  for machine learning,'' Tech. Rep., 2012.

\bibitem{Carreira_2017_CVPR}
J.~Carreira and A.~Zisserman, ``Quo vadis, action recognition? {A} new model
  and the kinetics dataset,'' in \emph{CVPR}, 2017.

\end{thebibliography}

\end{document}